\documentclass[10pt, conference, compsocconf]{IEEEtran}
\ifCLASSINFOpdf
\else
\fi
\usepackage{cite}
\usepackage{amsmath,amssymb,amsfonts}
\usepackage{algorithmic}
\usepackage{graphicx}
\usepackage{textcomp}
\usepackage{multirow}

\begin{document}
%
% paper title
% can use linebreaks \\ within to get better formatting as desired
\title{DenseRAN for Offline Handwritten Chinese Character Recognition}

% author names and affiliations
% use a multiple column layout for up to two different
% affiliations

\author{\IEEEauthorblockN{Wenchao Wang, Jianshu Zhang, Jun Du, Zi-Rui Wang and Yixing Zhu}
\IEEEauthorblockA{University of Science and Technology of China\\
Hefei, Anhui, P. R. China\\
Email: \{wangwenc, xysszjs\}@mail.ustc.edu.cn, jundu@ustc.edu.cn, \{cs211, zyxsa\}@mail.ustc.edu.cn}
}
% conference papers do not typically use \thanks and this command
% is locked out in conference mode. If really needed, such as for
% the acknowledgment of grants, issue a \IEEEoverridecommandlockouts
% after \documentclass

% for over three affiliations, or if they all won't fit within the width
% of the page, use this alternative format:
% 
%\author{\IEEEauthorblockN{Michael Shell\IEEEauthorrefmark{1},
%Homer Simpson\IEEEauthorrefmark{2},
%James Kirk\IEEEauthorrefmark{3}, 
%Montgomery Scott\IEEEauthorrefmark{3} and
%Eldon Tyrell\IEEEauthorrefmark{4}}
%\IEEEauthorblockA{\IEEEauthorrefmark{1}School of Electrical and Computer Engineering\\
%Georgia Institute of Technology,
%Atlanta, Georgia 30332--0250\\ Email: see http://www.michaelshell.org/contact.html}
%\IEEEauthorblockA{\IEEEauthorrefmark{2}Twentieth Century Fox, Springfield, USA\\
%Email: homer@thesimpsons.com}
%\IEEEauthorblockA{\IEEEauthorrefmark{3}Starfleet Academy, San Francisco, California 96678-2391\\
%Telephone: (800) 555--1212, Fax: (888) 555--1212}
%\IEEEauthorblockA{\IEEEauthorrefmark{4}Tyrell Inc., 123 Replicant Street, Los Angeles, California 90210--4321}}

% use for special paper notices
%\IEEEspecialpapernotice{(Invited Paper)}

% make the title area
\maketitle

\begin{abstract}
Recently, great success has been achieved in offline handwritten Chinese character recognition by using deep learning methods. Chinese characters are mainly logographic and consist of basic radicals, however, previous research mostly treated each Chinese character as a whole without explicitly considering its internal two-dimensional structure and radicals. In this study, we propose a novel radical analysis network with densely connected architecture (DenseRAN) to analyze Chinese character radicals and its two-dimensional structures simultaneously. DenseRAN first encodes input image to high-level visual features by employing DenseNet as an encoder. Then a decoder based on recurrent neural networks is employed, aiming at generating captions of Chinese characters by detecting radicals and two-dimensional structures through attention mechanism. The manner of treating a Chinese character as a composition of two-dimensional structures and radicals can reduce the size of vocabulary and enable DenseRAN to possess the capability of recognizing unseen Chinese character classes, only if the corresponding radicals have been seen in training set. Evaluated on ICDAR-2013 competition database, the proposed approach significantly outperforms whole-character modeling approach with a relative character error rate (CER) reduction of 18.54\%. Meanwhile, for the case of recognizing 3277 unseen Chinese characters in CASIA-HWDB1.2 database, DenseRAN can achieve a character accuracy of about 41\% while the traditional whole-character method has no capability to handle them.

\end{abstract}

\begin{IEEEkeywords}
radical analysis network, dense convolutional network, attention, offline handwritten Chinese character recognition

\end{IEEEkeywords}

% For peer review papers, you can put extra information on the cover
% page as needed:
% \ifCLASSOPTIONpeerreview
% \begin{center} \bfseries EDICS Category: 3-BBND \end{center}
% \fi
%
% For peerreview papers, this IEEEtran command inserts a page break and
% creates the second title. It will be ignored for other modes.
\IEEEpeerreviewmaketitle

\section{Introduction}
% no \IEEEPARstart
Handwritten Chinese characters recognition is a challenging problem due to the large number of character classes, confusion between similar characters, and distinct handwriting styles across individuals~\cite{b1,b2}. According to the type of data acquisition, handwriting recognition can be divided into online and offline. For offline handwritten Chinese characters recognition (HCCR), characters are gray-scaled images which are analyzed and classified into different classes. In traditional methods, the procedures for HCCR often include: image normalization, feature extraction, dimension reduction and classifier training. With the success of deep learning~\cite{b3}, convolutional neural network (CNN)~\cite{b4} has been applied successfully in this domain. The multi-column deep neural network (MCDNN)~\cite{b5} was the first CNN used for HCCR. The team from
Fujitsu used a CNN-based model to win the ICDAR-2013 HCCR competition~\cite{b6}. Zhong et al.~\cite{b7} improved the performance which outperforms the human performance. Li et al.~\cite{b8} from Fujitsu further improved the performance based on a single CNN model with augmented training data using distortion. The ensemble based methods can be further used to improve the performance with some tradeoff on speed and memory. Zhong et al.~\cite{b9} further improved the performance to by using spatial transformer network with residual network. However, these algorithms can only recognize Chinese characters appeared in training set and have no ability to recognize unseen Chinese characters. Moreover, these algorithms treat each Chinese character as a whole without considering the similarites and sub-structures among Chinese characters.  
\begin{figure}
	\centering
	\setlength{\abovecaptionskip}{0pt}
	\setlength{\belowcaptionskip}{0pt}
	\includegraphics[width=0.5\textwidth]{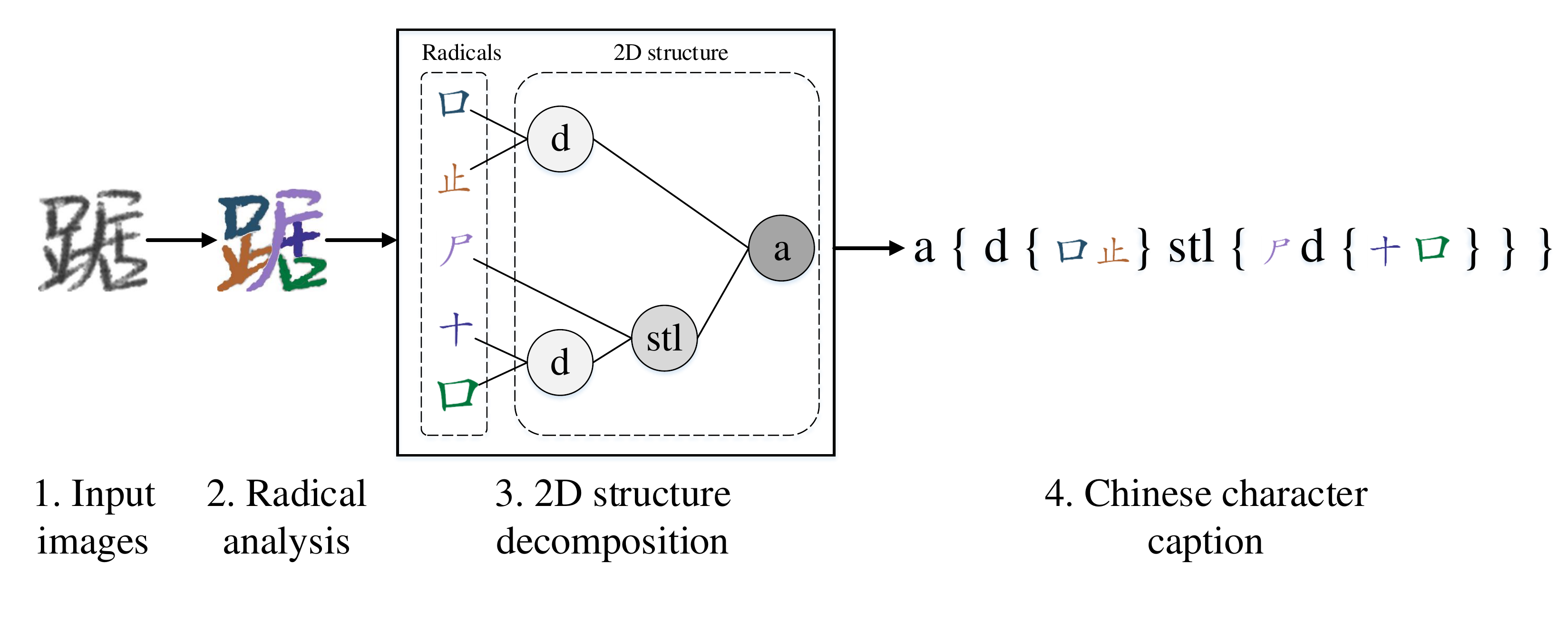}
	\caption{Illustration of DenseRAN to recognize Chinese characters by analyzing the radicals and two-dimensional structures. ``\textbf{d}" denotes top-bottom structure, ``\textbf{stl}" denotes top-left-surround structure, ``\textbf{a}" denotes left-right structure.}
	\label{fig:one}
\end{figure}

Chinese characters, which are mainly logographic and consist of basic radicals, constitute the oldest continuously used system of writing in the world and are different from the purely sound-based writing systems such as Greek, Hebrew, etc. It is natural to decompose Chinese characters to radicals and spatial structures then use this knowledge for character recognition. In the past few decades, a lot of work has been done for radical-based Chinese character recognition. \cite{b11} proposed a matching method which first detected radicals separately and then composed radicals into a character using a hierarchical radical matching method. \cite{b12} tried to over-segment characters into candidate radicals while the proposed way could only handle the left-right structure and over-segmentation brings many difficulties. Recently, \cite{b13} proposed a multi-label learning for radical-based Chinese character recognition. It turned a character class into a combination of several radicals and spatial structures. Generally, these approaches have difficulty in segmenting characters into radicals and lacking flexibility when to analyze structures among radicals. More importantly, they usually can't handle these unseen Chinese character classes.

In this paper, we propose a novel radical-based approach to HCCR, namely radical analysis network with densely connected architecture (DenseRAN). Different from above mentioned radical-based approaches, in DenseRAN the radical segmentation and structure detection are automatically learned by attention based encoder-decoder model. The main idea of DenseRAN is to decompose a Chinese character into a caption that describes its internal radicals and structures among radicals. A handwritten Chinese character is successfully recognized when its caption matches the groundtruth. In order to give a better explanation, we illustrate how DenseRAN recognizes a Chinese character in Fig.~\ref{fig:one}. Each leaf node of the tree in third step represents radicals and each non-leaf node represents its internel structure. The handwriting input is finally recognized as the Chinese character caption after the radicals and two-dimenstional structures are detected. Based on the analysis of radicals and structures, the proposed DenseRAN possesses the capability of recognizing unseen Chinese character classes only if the radicals have been seen in training set.

The proposed DenseRAN is an improved version of attention based encoder-decoder model~\cite{b14}. The overall architecture of DenseRAN is shown in Fig.~\ref{fig:three}. The raw data of %offline handwritten Chinese character 
input are gray-scaled images. DenseRAN first encodes input image to high-level visual vectors using a densely connected convolutional networks (DenseNet)~\cite{b22}. Then a RNN with gated recurrent units (GRU)~\cite{b23} decodes the high-level representations into output caption step by step. We adopt a coverage based spatial attention model built in the decoder to detect the radicals and internal two-dimensional structures simultaneously~\cite{b20,b21}. Compared with~\cite{b24} focusing on printed Chinese character recognition, DenseRAN focuses on HCCR which is much more difficult due to the diversity of writing styles.

%The main contributions of this paper are the follows:

%\begin{itemize}
%	\item
%	We propose DenseRAN for offline handwritten Chinese character recognition.
%	\item
%	We use CNN with densely connected architecture to improve the performance compared with VGG architecture~\cite{b24}.
%	\item
%	DenseRAN possess the ability of recognizing unseen or newly created Chinese characters, only if the radicals have been seen in training set.
%\end{itemize}
% You must have at least 2 lines in the paragraph with the drop letter
% (should never be an issue)

\section{Chinese character decomposition}
Each Chinese character can be naturally decomposed into a caption of radicals and spatial structures. Following the rule in\cite{b25}, the character caption consists three key components: radicals, spatial structures and a pair of braces (e.g. ``\{" and ``\}"). One spatial structure with its radicals can be represented as: ``\textbf{structure} \{ radical-1, radical-2 \}". 

\begin{figure}
	\centering
	\setlength{\abovecaptionskip}{0pt}
	\setlength{\belowcaptionskip}{0pt}
	\includegraphics[width=0.45\textwidth]{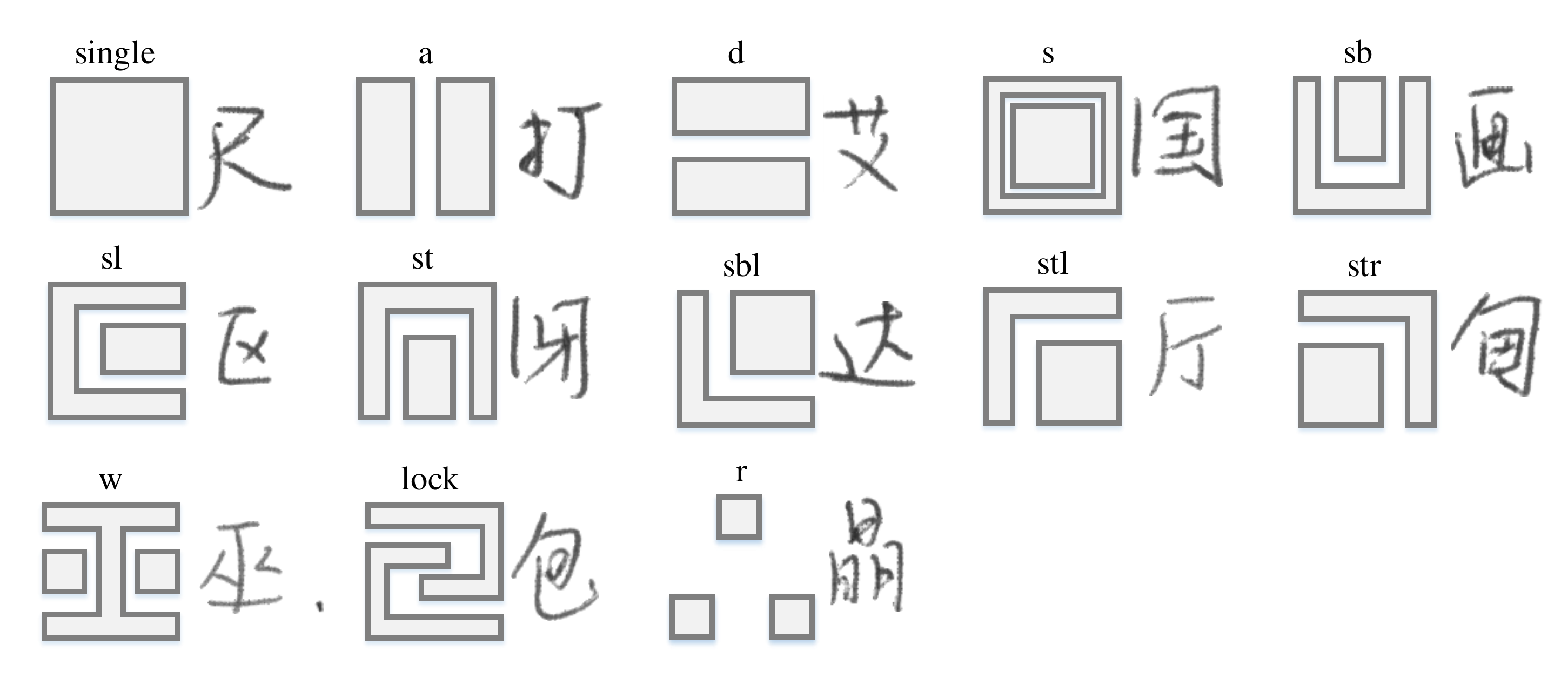}
	\caption{Graphical representation of thirteen common spatial structures.}% among Chinese characters.}
	\label{fig:two}
\end{figure}

A radical represents a basic part of Chinese character and is frequently shared among Chinese characters. Compared with enormous Chinese character categories, the total number of radicals is quite limited. It is declared in GB13000.1 standard published by National Language Committee of China that there are nearly 500 radicals for over 20,000 Chinese characters. Fig.~\ref{fig:two} illustrates thirteen common spatial structures and their corresponding Chinese character samples. These thirteen structures are: \textbf{single:} some Chinese characters are radicals themselves. \textbf{a:} left-right structure. \textbf{d:} top-bottom structure. \textbf{s:} surround structure. \textbf{sb:} bottom-surround structure. \textbf{sl:} left-surround structure. \textbf{st:} top-surround structure. \textbf{sbl:} bottom-left-surround structure. \textbf{stl:} top-left-surround structure. \textbf{str:} top-right-surround structure. \textbf{w:} within structure. \textbf{lock:} lock structure. \textbf{r:} one radical repeated many times in a character.

%\begin{itemize}
%	\item[-]
%	\textbf{single:} some Chinese characters are radicals themselves
%	\item[-]
%	\textbf{a:} left-right structure
%	\item[-]
%	\textbf{d:} top-bottom structure
%	\item[-]
%	\textbf{s:} surround structure
%	\item[-]
%	\textbf{sb:} bottom-surround structure
%	\item[-]
%	\textbf{sl:} left-surround structure
%	\item[-]
%	\textbf{st:} top-surround structure
%	\item[-]
%	\textbf{sbl:} bottom-left-surround structure
%	\item[-]
%	\textbf{stl:} top-left-surround structure
%	\item[-]
%	\textbf{str:} top-right-surround structure
%	\item[-]
%	\textbf{w:} within structure
%	\item[-]
%	\textbf{lock:} lock structure
%	\item[-]
%	\textbf{r:} one radical repeated many times in a character
%\end{itemize}

\section{The architecture of DenseRAN}
%DenseRAN first uses DenseNet and attention probabilities to encode input image into a fixed-length vector. Then a recurrent neural network uses this vector to generate the symbols of caption one by one. The overall architecture of DenseRAN is shown in Fig.~\ref{fig:three}.

\subsection{Dense encoder}\label{AA}

Dense convolutional network (DenseNet)~\cite{b22} has been proven to be good feature extractors for various computer vision tasks. So we use DenseNet as the encoder to extract high-level visual features from images. Instead of extracting features after a fully connected layer, we discard fully connected layer and softmax layer in encoder, called fully convolutional neural networks. This allows the decoder to selectively pay attention to certain parts of an image by choosing specific portions from the extracted visual features. 

\begin{figure}
	\centering
	\setlength{\abovecaptionskip}{0pt}
	\setlength{\belowcaptionskip}{0pt}
	\includegraphics[width=0.4\textwidth]{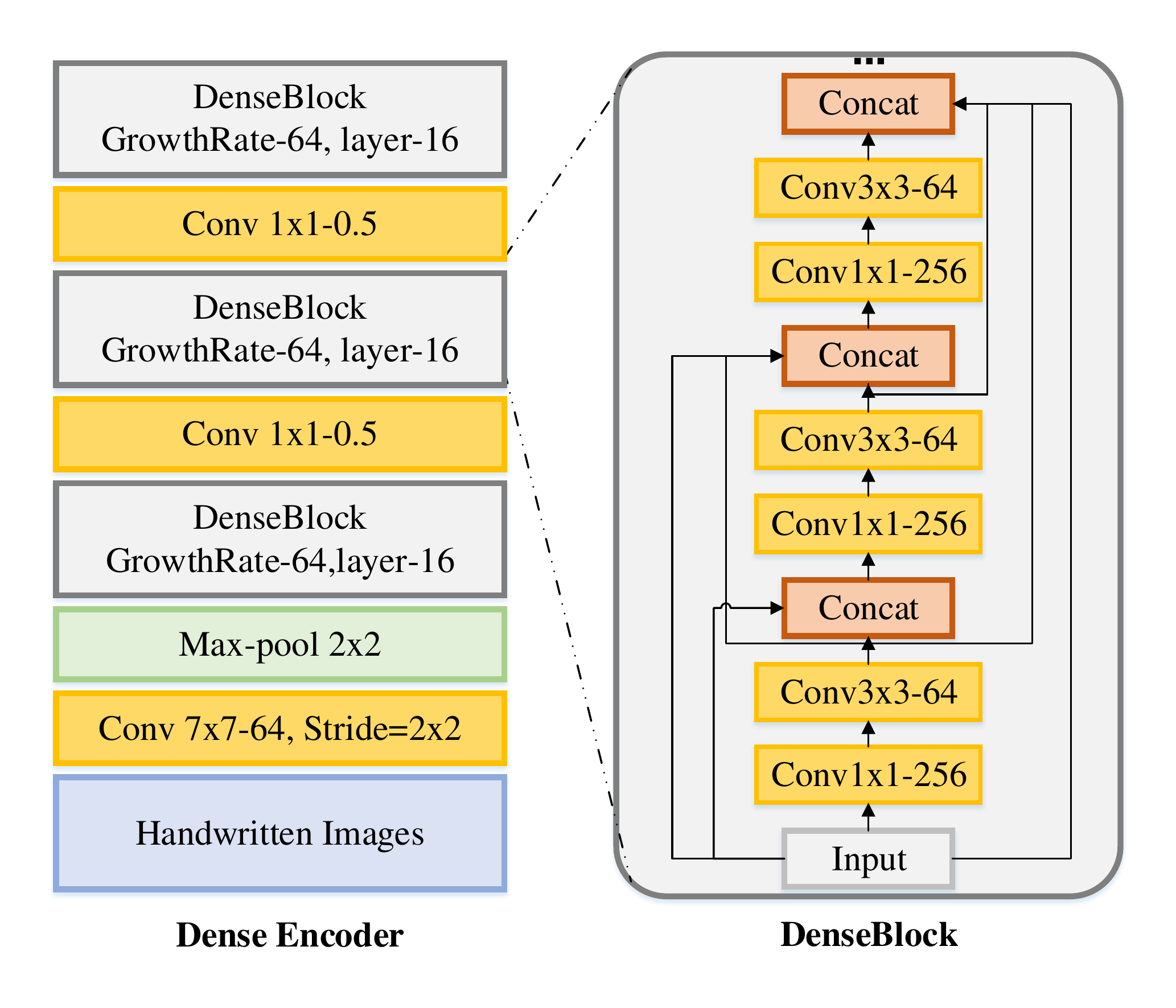}
	\caption{The architecture of DenseEncoder.}
	\label{fig:four}
\end{figure}

The architecture of DenseNet is mainly divided into multiple DenseBlocks. As shown in Fig.~\ref{fig:four}, in each denseblock, each layer is connected directly to all subsequent layers. We denote $H_l(\cdot)$ as the convolution function of the $l^{\text{th}}$ layer in some block, then the output of the $l^{\text{th}}$ layer can be represented as:

\begin{figure}
	\centering
	\setlength{\abovecaptionskip}{0pt}
	\setlength{\belowcaptionskip}{0pt}
	\includegraphics[width=0.45\textwidth]{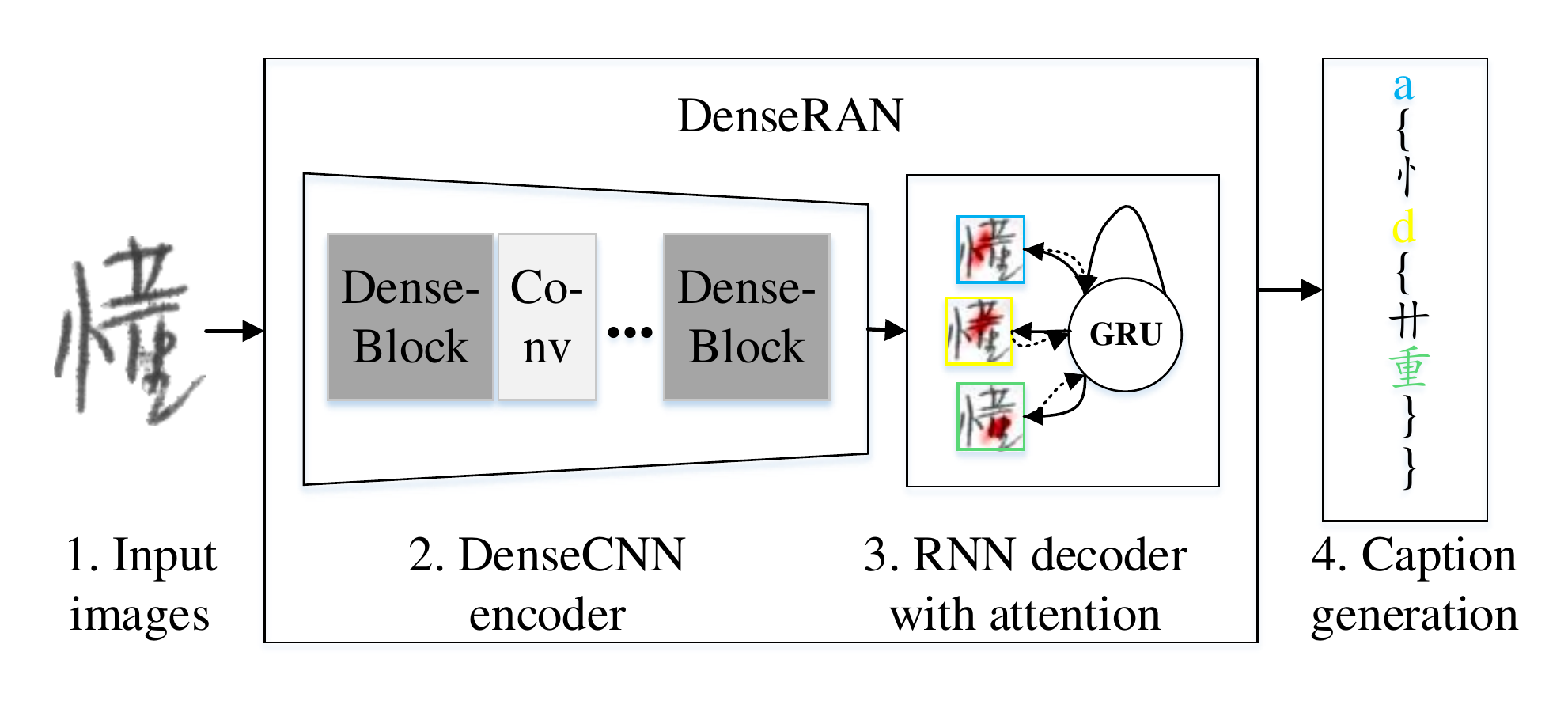}
	\caption{The overall architecture of DenseRAN for HCCR.}
	\label{fig:three}
\end{figure}
\begin{equation}
\text{x}_l = H_l([\text{x}_0,\text{x}_1,\ldots,\text{x}_{l-1}])
\end{equation}
where $[\text{x}_0,\text{x}_1,\ldots,\text{x}_{l-1}]$ denotes the concatenation of the output feature maps produced by $0,1\ldots,l-1$ in the same block. The growth rate $k=64$ means each $H_l(\cdot)$ produces $k$ feature maps. In order to further improve computational efficiency, we use bottleneck layers in each DenseBlock. A $1\times1$ convolution is introduced before each $3\times3$ convolution to reduce the number of feature maps input to $4k$. The depth of each Denseblock is set to $D=16$, i.e., in each block, there are $D$ $1\times1$ convolution layers and each one is followed by a $3\times3$ convolution layer. 

These DenseBlocks are normally seperated by transition layer and pooling layer. Each transition layer is a $1\times1$ convolution layer parameterized by $\theta=0.5$. If the number of input feature maps of transition layer is $n$, then the transition layer will generate $\theta n$ output feature maps. In DenseRAN, because the input character images are resized to $32\times32$, after so many pooling operation, the size of final feature map is about $2\times2$, which is too small to get good attention results. So we discard the pooling layer between DenseBlocks in Fig.~\ref{fig:four}. The first convolution layer has $64$ convolutions of kernel size $7\times7$ with stride $2$ which is performed on the input images, followed by a $2\times2$ max pooling layer. Batch normalization\cite{b26} and ReLU\cite{b27} are performed after each convolution layer consecutively.

Dense encoder extracts visual features which can be represented as a three-dimensional array of size $H \times W \times D$, $L=H \times W$. Each element in array is a $D$-dimensional vector that corresponds to a local region of the image:
\begin{equation}
\textbf{A} = \{\textbf{a}_1,\ldots,\textbf{a}_L\}, \textbf{a}_i \in \mathbb{R}^D
\end{equation}

\subsection{GRU decoder with attention model}
As illustrated in Fig.~\ref{fig:three}, the decoder generates a caption of input Chinese character. The output caption $\textbf{Y}$ is represented by a sequence of 1-of-$K$ encoded symbols:
\begin{equation}
\textbf{Y} = \{\textbf{y}_1,\ldots,\textbf{y}_C\}, \textbf{y}_i \in \mathbb{R}^K
\end{equation}
where $K$ is the number of total symbols in the vocabulary which includes the basic radicals, spatial structures and a pair of braces, $C$ is the length of caption.

Because the length of annotation sequence $L$ is fixed while the length of captions $C$ is variable, DenseRAN addresses this problem by computing an intermediate fixed-size vector $\textbf{c}_t$ at each time step. Note that $\textbf{c}_t$ is a dynamic representation of the relevant part of the Chinese character image at time $t$. We utilize unidirectional GRU\cite{b28} and the context vector $\textbf{c}_t$ to produce captions step by step. %The GRU is an improved version of simple RNN as it alleviates the problems of the vanishing gradient and the exploding gradient as described in\cite{b28}. 
The probability of each predicted word is computed by the context vector $\textbf{c}_t$, current GRU hidden state $\textbf{s}_t$ and previous word $\textbf{y}_{t-1}$ using the following equation:
\begin{equation}
p(\textbf{y}_t|\textbf{y}_{t-1},\textbf{X}) = \text{Softmax}(\textbf{W}_0(\textbf{Ey}_{t-1}+\textbf{W}_s\textbf{s}_t+\textbf{W}_c\textbf{c}_t))
\end{equation}
where $\textbf{W}_0 \in \mathbb{R} ^ {K \times m}$, $\textbf{W}_s \in \mathbb{R} ^ {m \times n}$, $\textbf{W}_c \in \mathbb{R} ^ {m \times D}$, and $\mathbf{E}$ denotes the embedding matrix, $m$ and $n$ are the dimensions of embedding and GRU parser.

The GRU parser adopts two unidirectional GRU layers to calculate the hidden state $\textbf{s}_t$:
\begin{eqnarray}
\hat{\textbf{s}}_t &=& \text{GRU}(\textbf{y}_{t-1},\textbf{s}_{t-1})\\
\textbf{c}_t &=& f_{\text{att}}(\hat{\textbf{s}}_t,\textbf{A})\\
\textbf{s}_t &=& \text{GRU}(\hat{\textbf{s}}_t,\textbf{c}_t)
\end{eqnarray}
where $\textbf{s}_{t-1}$ denotes hidden state at time $t-1$, $\hat{\textbf{s}}_t$ is the GRU hidden state prediction at time $t$, and coverage based spatial attention model $f_{\text{att}}$ is parameterized as a multi-layer perceptron:
\begin{eqnarray}
\textbf{F} &=& \textbf{Q} * \sum_{l=1}^{t-1} \boldsymbol{\alpha}_l \\
e_{ti} &=& \textbf{v}_{\text{att}}^T \text{tanh}(\textbf{W}_{\text{att}}\hat{\textbf{s}}_t+\textbf{U}_{\text{att}}\textbf{a}_i + \textbf{U}_f \textbf{f}_i) \\
\alpha_{ti} &=& \frac {\exp(e_{ti})} {\sum_{k=1}^L \exp(e_{tk})}
\end{eqnarray}

%where $e_{ti}$ denotes the energy of annotation vector $\textbf{a}_i$ at time step $t$ condition on the current GRU hidden state prediction $\hat{\textbf{s}}_t$ and coverage vector $\textbf{f}_i$. 
The coverage vector $\textbf{F}$ %is initialized as a zero vector and we compute it based on the summation of past attention probabilities. 
is computed based on the summation of past attention probabilities. $\alpha_{ti}$ denotes the spatial attention coefficient of $\textbf{a}_i$ at time $t$. Let $n'$ denotes the attention dimension and $q$ denotes the feature map of filter $\textbf{Q}$, then $\textbf{v}_{\text{att}} \in \mathbb{R} ^ {n'}$, $\textbf{W}_{\text{att}} \in \mathbb{R} ^ {n' \times n}$, $\textbf{U}_{\text{att}} \in \mathbb{R} ^ {n' \times D }$, $\textbf{U}_f \in \mathbb{R} ^ {n' \times q}$. With the weight $\alpha_{ti}$, we compute the context vector $\textbf{c}_t$ as:
\begin{equation}
\textbf{c}_t = \sum_{i=1}^L \alpha_{ti} \textbf{a}_i
\end{equation}

\section{Experiments on recognizing seen Chinese characters}

%In this section, we present experiments on seen offline Chinese characters, for the perpose of evaluating and comparing DenseRAN with traditional approaches.
In this section, we present some comparison experiments on seen offline Chinese characters to show the advantage of performance of DenseRAN.

\subsection{Dataset}

The database used for evaluation is from the ICDAR-2013 competition~\cite{b6} of HCCR. The database used for training is the CASIA database~\cite{b29} including HWDB1.0 and 1.1. The most common Chinese characters are used, i.e., 3755 level-1 set of GB2312-80.

\subsection{Implementation details}

%In this paper, each character is represented by a grayscaled image. 
We normalize gray-scaled image to the size of $32 \times 32$ as the input. The implementation details of Dense encoder has been introduced in Section III-A. The decoder is two unidirectional layers with $256$ GRU units. The embedding dimension $m$ and decoder state dimension $n$ are set to $256$. The convolution kernel of $\textbf{Q}$ is set to $5 \times 5$ and the number of feature maps is set to $128$. The model is trained with mini-batch size of $150$ on one GPU. We utilize the adadelta \cite{b30} with gradient clipping for optimization. The best model is determined in terms of word error rate (WER) of validation set. We use a weight decay of $10^{\textnormal{-}4}$ and dropout\cite{b31} after each convolution layer and set the dropout rate to $0.2$.

In the decoding stage, we aim to generate a most likely caption string given the input character. The beam search algorithm \cite{b32} is employed to find the optimal decoding path in the decoding process. %because we do not have the ground-truth of previous predicted word during testing.
The beam size is set to $10$.

\subsection{Experiments results}

\begin{table}[t]
	\begin{center}
		%\caption{Results on ICDAR-2013 competition database of offline handwritten Chinese Character recognition.} 
		\caption{Results on ICDAR-2013 competition database of HCCR.}
		\label{tab:two}
		\newcommand{\tabincell}[2]{\begin{tabular}{@{}#1@{}}#2\end{tabular}} 
		\begin{tabular}{|c|c|c|}
			\hline
			% after \\: \hline or \cline{col1-col2} \cline{col3-col4} ...
			Methods & Ref. & Accuracy
			\\
			\hline
			Human Performance & \cite{b6} & 96.13\%\\
			\hline
			Traditional Method &\cite{b33} & 92.72\%\\
			\hline
			VGG14RAN & - & 93.79\%\\
			\hline
			DenseNet & - & 95.90\%\\
			\hline
			DenseRAN & - &  96.66\%\\
			\hline
		\end{tabular}
	\end{center}
\end{table}

In Table \ref{tab:two}, the human performance on ICDAR-2013 competition database and the previous benchmark are both listed. In order to compare DenseRAN with whole-character based approach, only DenseNet which is the same as the encoder of DenseRAN is evaluated as a whole-character classifier on ICDAR-2013 competition database, we call it ``DenseNet". As shown in Table \ref{tab:two}, ``DenseNet" achieves 95.90\% while DenseRAN achieves 96.66\% revealing relative character error rate reduction of 18.54\%. Also, we replace the encoder of DenseRAN with VGG14\cite{b34} and keep the other parts unchanged, we name it as ``VGG14RAN". Table \ref{tab:two} clearly shows CNN with densely connnected architecture is more powerful than VGG on extracting high-quality visual features from handwritten Chinese character images.

\section{Experiments on recognizing unseen Chinese characters}

Chinese characters are enormous which is difficult to train a recognition system that covers all of them. %Therefore it is necessary for a system to possess the capability of recognizing unseen Chinese characters. In this section, we show the effectiveness of DenseRAN to identify unseen characters. 
Therefore it is necessary and interesting to empower a system to recognize unseen Chinese characters. In this section, we show the effectiveness of DenseRAN to recognize unseen characters.

\subsection{Dataset}
We divide 3755 common Chinese characters into 2755 classes and another 1000 classes. %We pick handwritten characters belonging to 2755 classes from HWDB1.0 and 1.1 set as new training set and handwritten characters belonging to the other 1000 classes from ICDAR-2013 competition set as new test set. 
We pick 2755 classes in HWDB1.0 and 1.0 set as the training set and the other 1000 classes in ICDAR-2013 competition database are selected as the test set. So both the test character classes and handwriting styles have never been seen during training. In order to explore the influence of training samples, different training sizes are designed, ranging from 500 to 2755 classes. Note that all radicals are covered in all training sets. %We explore different size of training set to train DenseRAN, ranging from 500 to 2755 classes and we also make sure the radicals of test characters are all covered in training set.

HWDB1.2 dataset is also used for evaluating the DenseRAN performance on unseen Chinese characters. There are 3319 non-common Chinese characters in HWDB1.2 dataset and we pick 3277 classes to make sure the radicals of these characters are covered in 3755 common classes. Note that the Chinese characters in HWDB1.2 dataset are not common and usually have more complicated radical composition.

\subsection{Experiments results}

\begin{table}[t]
	\begin{center}
		%\caption{Results on unseen offline handwritten Chinese Character recognition.} 
		\caption{Results on unseen HCCR.} 
		\label{tab:three}
		\newcommand{\tabincell}[2]{\begin{tabular}{@{}#1@{}}#2\end{tabular}} 
		\begin{tabular}{|c|c|c|c|c|c|c|}
			\hline
			% after \\: \hline or \cline{col1-col2} \cline{col3-col4} ...
			\tabincell{c}{Train \\ set} & \tabincell{c}{Train \\ class} & \tabincell{c}{Train \\ samples} & \tabincell{c}{Test\\ set} & \tabincell{c}{Test \\ class}  & Accuracy
			%\tabincell{c}{Train set} & \tabincell{c}{Train class} & \tabincell{c}{\#Train} & \tabincell{c}{Test set} & \tabincell{c}{Test class}  & Accuracy
			\\
			\hline
			\multirow{5} * {\tabincell{c}{HWDB\\1.0+1.1}} & 500 & 356553 & \multirow{5} * {\tabincell{c}{ICDAR\\2013}} & 1000 & 1.70\%\\
			\cline{2-3} \cline{5-6}
			~ & 1000 & 712254 & ~ & 1000 & 8.44\%\\
			\cline{2-3} \cline{5-6}
			~ & 1500 & 1068031 & ~ & 1000 & 14.71\%\\
			\cline{2-3} \cline{5-6}
			~ & 2000 & 1425530 & ~ & 1000 & 19.51\%\\
			\cline{2-3} \cline{5-6}
			~ & 2755 & 1962529 & ~ & 1000 & 30.68\%\\
			\hline
			\tabincell{c}{HWDB\\1.0+1.1} & 3755 & 2674784 & \tabincell{c}{HWDB\\1.2} & 3277 & 40.82\%\\
			\hline
		\end{tabular}
	\end{center}
\end{table}

As shown in Table \ref{tab:three}, with the seen Chinese character classes increase from 500 to 2755, the accuracy on 1000-class test set increases from 1.70\% to 30.68\%. Whole-character modeling systems can not recognize unseen Chinese character classes at all. %which means their accuracies are definitely \textbf{0\%}. 
The last row of Table \ref{tab:three} shows that DenseRAN can recognize unseen uncommon Chinese characters in HWDB1.2 with 40.82\% accuracy.

\section{Qualitative Analysis}

\subsection{Attention visualization}

\begin{figure}
	\centering
	\setlength{\abovecaptionskip}{0pt}
	\setlength{\belowcaptionskip}{0pt}
	\includegraphics[width=0.45\textwidth]{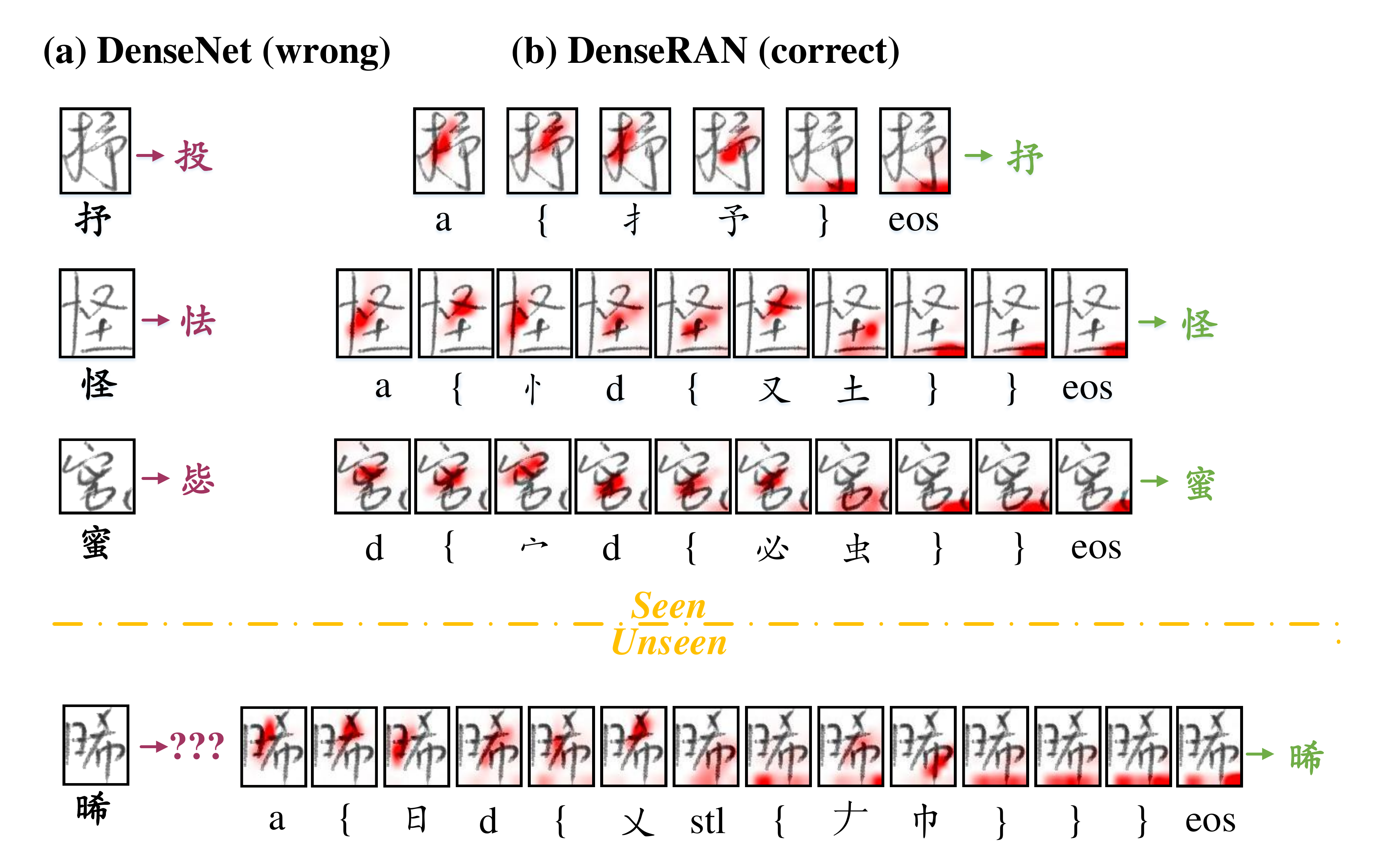}
	\caption{(a): Characters misclassified by ``DenseNet". (b): Same characters classified correctly by DenseRAN and the attention over times.}
	\label{fig:five}
\end{figure}

By visualizing the attention probabilities learned by the model, we show how DenseRAN recognizes radicals and two-dimensional structures. We also analyze the error examples by attention visualization. We show some Chinese chracters which are misclassified by ``DenseNet" in Fig.~\ref{fig:five}(a). On the contrary, as shown in Fig.~\ref{fig:five}(b), %DenseRAN learns alignments for offline handwritten Chinese character as human intuition and get the correct classification. 
DenseRAN aligns radicals and structures of offline handwritten Chinese character step by step as human intuition and finally gets the correct classification. Above the dotted line, these Chinese characters are seen in training set. Below the dotted line, the character is not seen in training set. Fig.~\ref{fig:five} clearly illustates that DenseRAN not only outperforms whole-character modeling method, but also has the ability of recognizing unseen characters.

\begin{figure}
	\centering
	\setlength{\abovecaptionskip}{0pt}
	\setlength{\belowcaptionskip}{0pt}
	\includegraphics[width=0.5\textwidth]{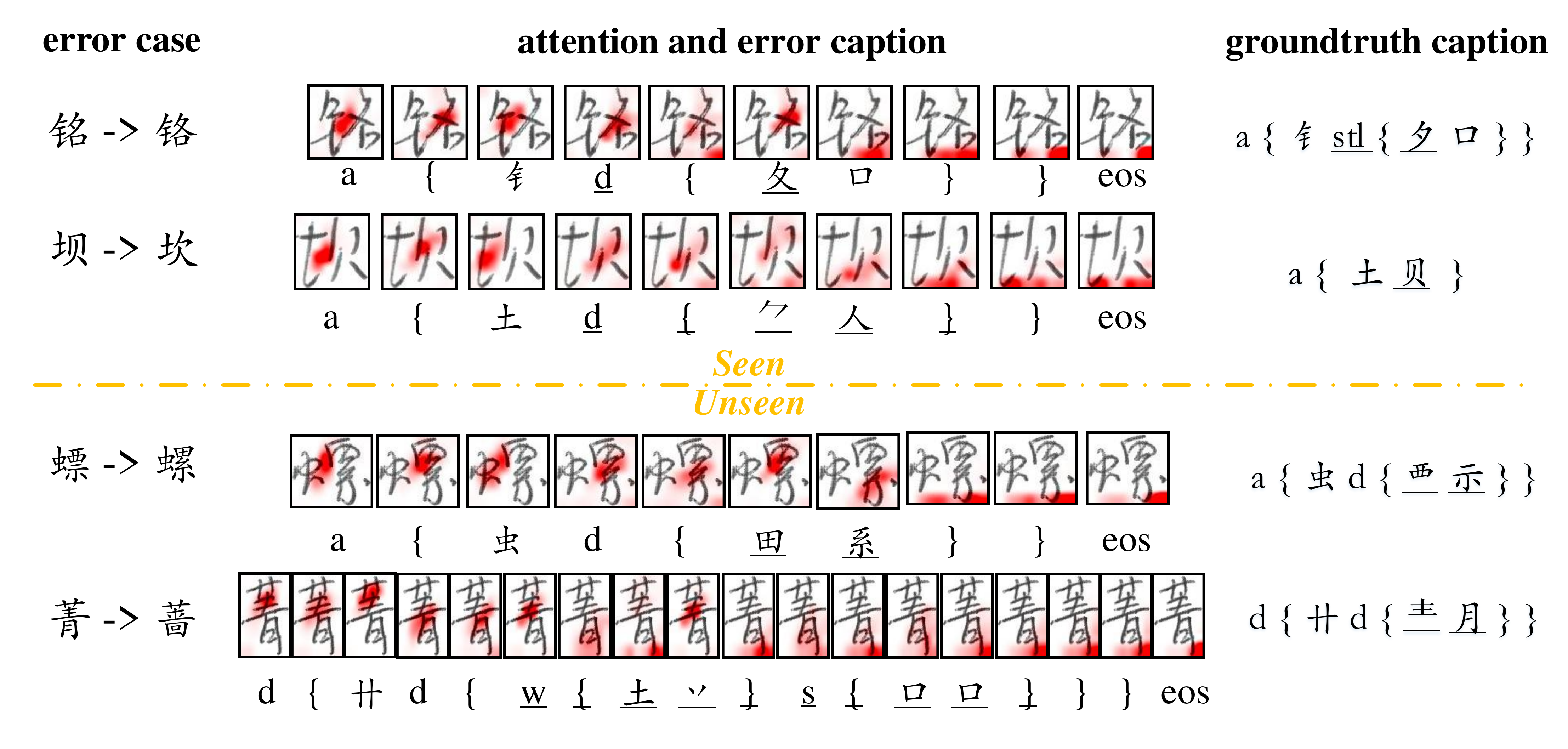}
	\caption{Examples of mistakes where we can use attention to gain intuition into what DenseRAN saw.}
	\label{fig:six}
\end{figure}

Examples of mistakes are shown in Fig.~\ref{fig:six}. The first column shows the correct characters and the misclassified characters as well. The second column visualizes attention and the symbols decoded over times. The third column gives the corresponding groundtrurth caption. Above the dotted line, these Chinese characters are seen in training set. Below the dotted line, these characters are not seen in training set. As shown in Fig.~\ref{fig:six}, the mistaken characters usually come from the confuable characters which are easy to cause errors when the handwritten style is scribbled. 

%This is why HCCR is more difficult than printed Chinese character recognition (PCCR), and also this is why the performance of DenseRAN on recognizing unseen Chinese character is not as good as \cite{b24}, which is implemented on PCCR.

\subsection{Error distribution of different two-dimensional structures}
\begin{figure}
	\centering
	\setlength{\abovecaptionskip}{0pt}
	\setlength{\belowcaptionskip}{0pt}
	\includegraphics[width=0.4\textwidth]{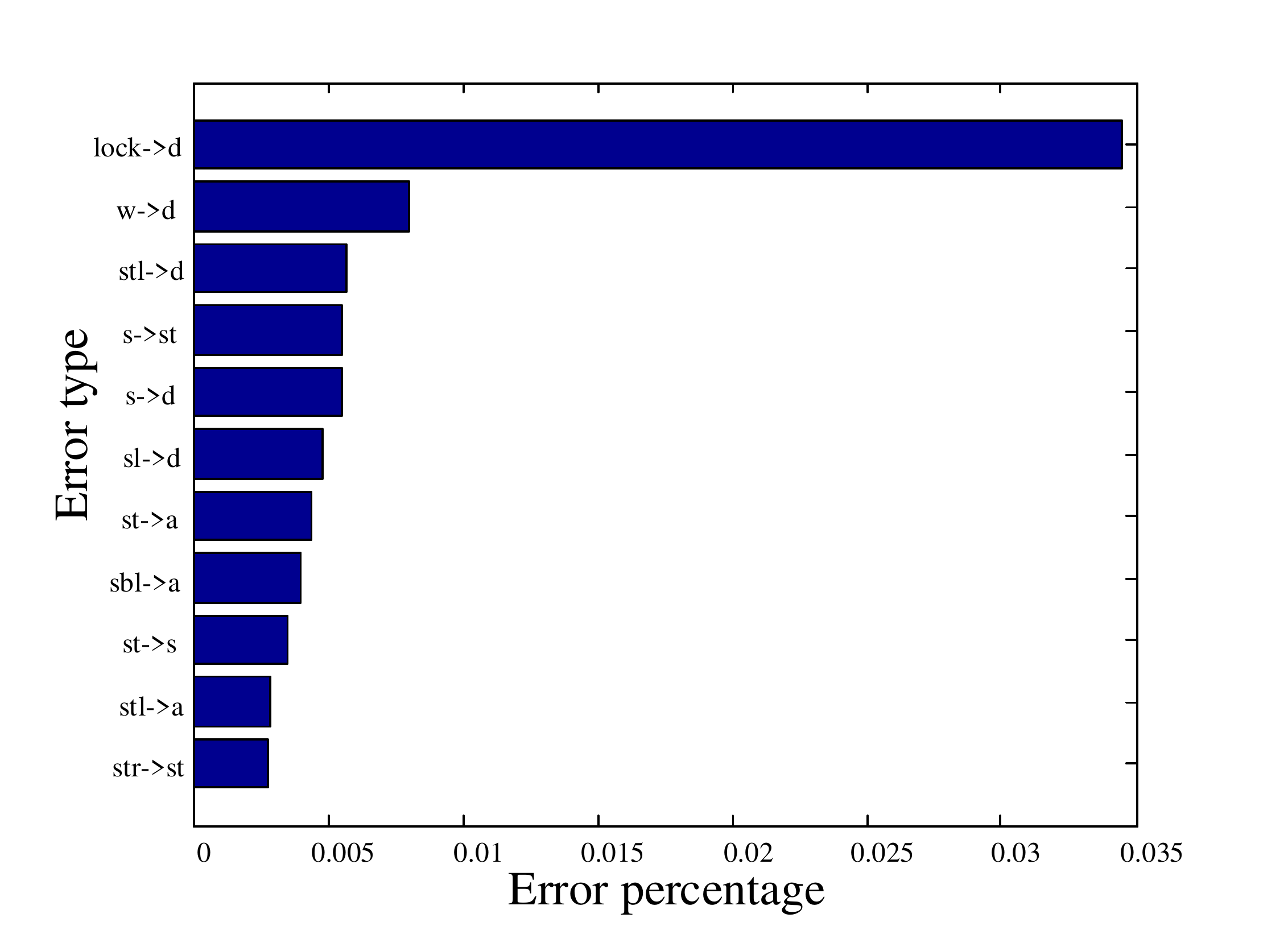}
	\caption{Error percentage of different two dimensional structures.}
	\label{fig:seven}
\end{figure}

\begin{figure}
	\centering
	\setlength{\abovecaptionskip}{0pt}
	\setlength{\belowcaptionskip}{0pt}
	\includegraphics[width=0.4\textwidth]{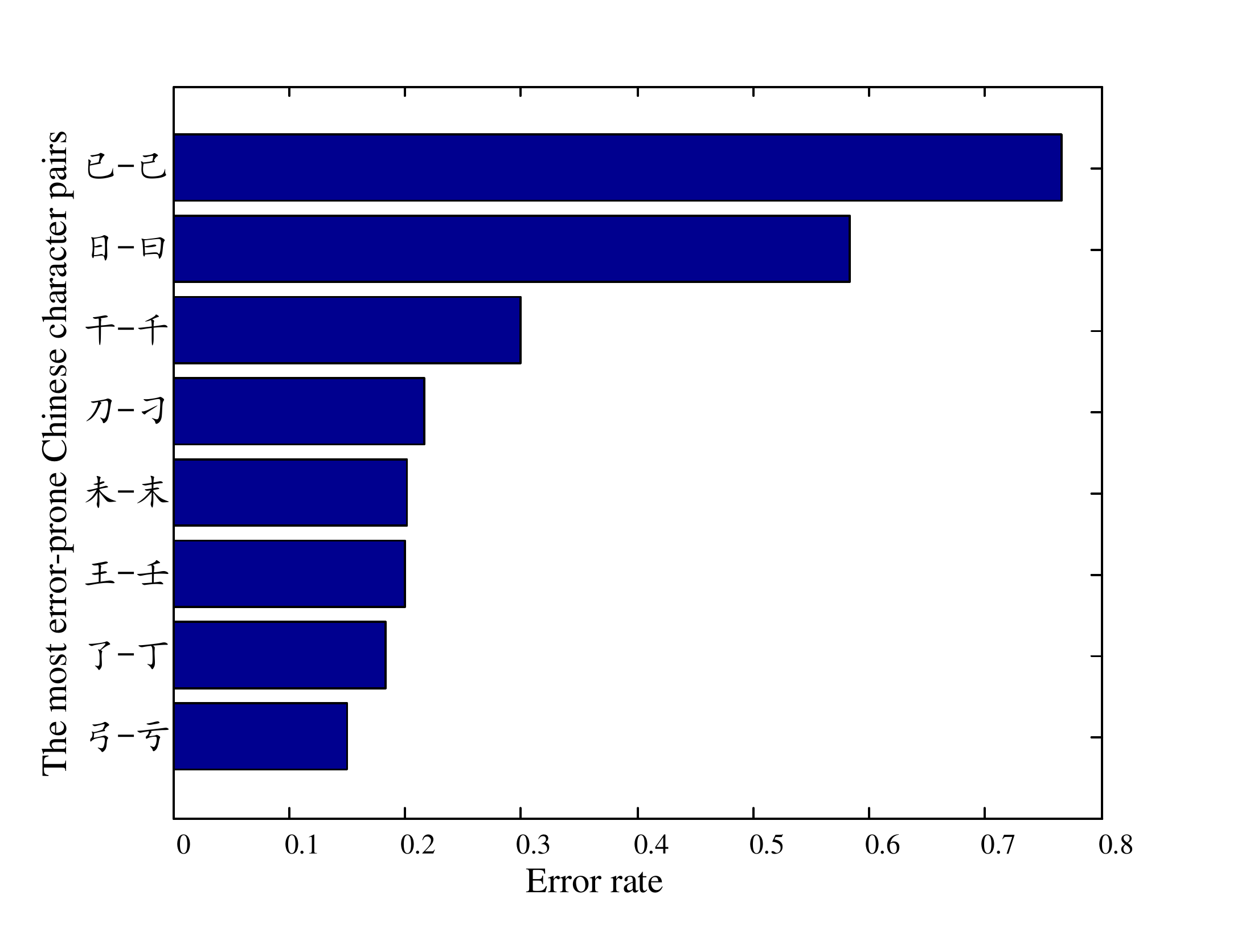}
	\caption{Error rate of Chinese characters with \textbf{single} structure.}
	\label{fig:eight}
\end{figure}
In order to further explore important factors that cause the error, we analyze the error distribution with different two-dimensional structures of Chines characters. Intuitively, %the more complicated the structure, the more error-prone.
the more complicated the structure is, the more easily the error occurs. Fig.~\ref{fig:seven} shows the error percentage of structure $A$ misclassified to structure $B$ on test set except the ``\textbf{single}" structure. The error percentage here is calculated as:
\begin{eqnarray}
P_{err}(A \rightarrow B) = \frac {N_{A \rightarrow B}} {N_A}
\end{eqnarray}
where ${N_A}$ denotes the number of samples with structure $A$ in test set, $N_{A \rightarrow B}$ denotes the number of samples with structure $A$ misclassified to $B$. As shown in Fig.~\ref{fig:seven}, the most likely mistake is ``$\textbf{lock}$" misclassified to ``$\textbf{d}$" with 3.45\% error percentage. The explanation is: as shown in Fig.~\ref{fig:two}, ``$\textbf{lock}$" will become ``$\textbf{d}$" when the two parts of ``$\textbf{lock}$" are written seperately. This situation is very easy to appear when the handwritten style is scribbled. 

Fig.~\ref{fig:eight} shows the error rate of error pair ($E$-$G$) which are both ``\textbf{single}" structure, which is calculated as:
\begin{eqnarray}
R_{err}(E\text{-}G) = \frac {N_{E \rightarrow G} + N_{G \rightarrow E}} {N_E + N_G} 
\end{eqnarray}
where $N_E$ denotes the number of character $E$ in test set, $N_G$ denotes the number of character $G$. $N_{E \rightarrow G}$ denotes how many $E$ samples are misclassified to $G$ in test set, vice versa. From Fig.~\ref{fig:eight}, we find the misclassified characters usually come from the characters which only have some subtle differences in radicals with another character.

\section{Conclusion and Future work}
In this study we introduce DenseRAN for %offline handwritten Chinese character recognition
HCCR. The proposed DenseRAN recognizes Chinese character by identifying radicals and analyzing spatial structures. Experiments shows that DenseRAN outperforms whole-character modeling approach on HCCR task and has the capability of recognizing unseen Chinese characters. By visualizing attention and analyzing error examples, we should pay more attention on confusing characters in the future.

% conference papers do not normally have an appendix

% use section* for acknowledgement
\section*{Acknowledgment}
This work was supported in part by the National Key R\&D Program of China under contract No. 2017YFB1002202, the National Natural Science Foundation of China under Grants No. 61671422 and U1613211, the Key Science and Technology Project of Anhui Province under Grant No. 17030901005, and MOE-Microsoft Key Laboratory of USTC. This work was also funded by Huawei Noah's Ark Lab.

% trigger a \newpage just before the given reference
% number - used to balance the columns on the last page
% adjust value as needed - may need to be readjusted if
% the document is modified later
%\IEEEtriggeratref{8}
% The "triggered" command can be changed if desired:
%\IEEEtriggercmd{\enlargethispage{-5in}}

% references section

% can use a bibliography generated by BibTeX as a .bbl file
% BibTeX documentation can be easily obtained at:
% http://www.ctan.org/tex-archive/biblio/bibtex/contrib/doc/
% The IEEEtran BibTeX style support page is at:
% http://www.michaelshell.org/tex/ieeetran/bibtex/
%\bibliographystyle{IEEEtran}
% argument is your BibTeX string definitions and bibliography database(s)
%\bibliography{IEEEabrv,../bib/paper}

\begin{thebibliography}{1}

\bibitem{b1} F. Kimura, K. Takashina, S. Tsuruoka, Y. Miyake, ``Modified quadratic discriminant functions and the application to Chinese character recognition," \emph{IEEE Trans. Pattern Anal. Mach. Intell.}, pp. 149--153, 1987.

\bibitem{b2} R.-W. Dai, C.-L. Liu, B.-H. Xiao, ``Chinese character recognition: history, status and prospects," \emph{Front. Comput. Sci. China}, pp. 126--136, 2007.

\bibitem{b3} LeCun, Y. and Bengio, Y. and Hinton, G., ``Deep learning," \emph{Nature}, vol. 521, no. 7553, pp. 436--444, 2015.

\bibitem{b4} LeCun, Y. and Bottou, L. and Bengio, Y. and Haffner, P., ``Gradient-based learning applied to document recognition," \emph{Proceedings of the IEEE}, vol. 86, no. 11, pp. 2278--2324, 1998.

\bibitem{b5} D. Ciresan, J. Schmidhuber, ``Multi-column deep neural networks for offline handwritten Chinese character classification," \emph{arXiv}:1309.0261, 2013.

\bibitem{b6} F. Yin, Q.-F. Wang, X.-Y. Zhang, and C.-L. Liu, ``ICDAR 2013 Chinese handwriting recognition competition," \emph{Proc. ICDAR}, 2013, pp. 1464--1470.

\bibitem{b7} Z. Zhong, L. Jin, and Z. Xie, ``High performance offline handwritten Chinese character recognition using GoogLeNet and directional feature maps," \emph{Proc. ICDAR}, 2015, pp. 846--850.

\bibitem{b8} L. Chen, S. Wang, W. Fan, J. Sun, and N. Satoshi, ``Beyond human recognition: A CNN-Based framework for handwritten character recognition," \emph{Proc. ACPR}, 2015, pp. 695--699.

\bibitem{b9} Z. Zhong, X-Y. Zhang, F. Yin, C.-L. Liu, ``Handwritten Chinese character recognition with spatial transformer and deep residual networks," \emph{Proc. ICPR}, 2016, pp. 3440--3445.

%\bibitem{b10} D. Jurafsky and J. Martin., ``\emph{Speech and language processing,}" vol. 3, Pearson London, 2014.

\bibitem{b11} A.-B. Wang and K.-C. Fan, ``Optical recognition of handwritten Chinese characters by hierarchical radical matching method," \emph{Pattern Recognition}, vol. 34, no. 1, pp. 15–-35, 2001.

\bibitem{b12} L.-L. Ma and C.-L. Liu, ``A new radical-based approach to online handwritten Chinese character recognition," in \emph{Pattern Recognition}, 2008. \emph{ICPR} 2008. \emph{19th International Conference on. IEEE}, 2008, pp. 1–-4.

\bibitem{b13} T.-Q. Wang, F. Yin, and C.-L. Liu, ``Radical-based Chinese character recognition via multi-labeled learning of deep residual networks," in \emph{Proc. ICDAR}, 2017, pp. 579--584.

\bibitem{b14} D. Bahdanau, K. Cho, and Y. Bengio, ``Neural machine translation by jointly learning to align and translate," \emph{arXiv}:1409.0473, 2014.

\bibitem{b20} J. Zhang, J. Du, S. Zhang, D. Liu, Y. Hu, J. Hu, S. Wei, and L. Dai, ``Watch, attend and parse: An end-to-end neural network based approach to handwritten mathematical expression recognition," \emph{Pattern Recognition}, 2017.

\bibitem{b21} J. Zhang, J. Du, and L. Dai, ``Multi-scale attention with dense encoder for handwritten mathematical expression recognition," \emph{arXiv}:1801.03530, 2018.

\bibitem{b22} Gao Huang, Zhuang Liu, Laurens van der Maaten, Kilian Q. Weinberger, ``Densely Connected Convolutional Networks," \emph{arXiv}:1608.06993, 2016.

\bibitem{b23} Junyoung Chung, Caglar Gulcehre, KyungHyun Cho, and Yoshua Bengio, ``Empirical evaluation of gated recurrent neural networks on sequence modeling," \emph{arXiv}:1412.3555, 2014.

\bibitem{b24} J. Zhang, Y. Zhu, J. Du, and L. Dai, ``Radical analysis network for zero-shot learning in printed Chinese character recognition," in \emph{Proc. International Conference on Multimedia and Expo}, 2018.

\bibitem{b25} A. Madlon-Kay, ``\emph{cjk-decomp.}" Available: https://github.com/amake/cjk-decomp.

\bibitem{b26} S. Ioffe and C. Szegedy, ``Batch normalization: Accelerating deep network training by reducing internal covariate shift," in \emph{Proc. 32nd International Conference on Machine Learning}, 2015, pp. 448–-456.

\bibitem{b27} V. Nair and G. E. Hinton, ``Rectified linear units improve restricted boltzmann machines," in \emph{Proc. 27th International Conference on Machine Learning}, 2010, pp. 807–-814.

\bibitem{b28} Yoshua Bengio, Patrice Simard, and Paolo Frasconi, ``Learning long-term dependencies with gradient descent is difficult," \emph{IEEE transactions on neural networks}, vol. 5, no. 2, pp. 157–-166, 1994.

\bibitem{b29} C.-L. Liu, F. Yin, D.-H. Wang, and Q.-F. Wang, ``CASIA online and offline Chinese handwriting databases," \emph{Proc. ICDAR}, 2011.

\bibitem{b30} M. D. Zeiler, ``ADADELTA: an adaptive learning rate method," \emph{arXiv}:1212.5701, 2012.

\bibitem{b31} N. Srivastava, G. Hinton, A. Krizhevsky, I. Sutskever, and R. Salakhutdinov, ``Dropout: A simple way to prevent neural networks from overfitting," \emph{Journal of Machine Learning Research}, vol. 15, no. 1, pp. 1929–-1958, 2014.

\bibitem{b32} K. Cho, ``Natural language understanding with distributed representation," \emph{arXiv}:1511.07916, 2015.

\bibitem{b33} C.-L. Liu, F. Yin, D.-H. Wang, and Q.-F. Wang, ``Online and offline handwritten Chinese character recognition: benchmarking on new databases," \emph{Pattern Recognition}, vol. 46, no. 1, pp. 155–-162, 2013.

\bibitem{b34} K. Simonyan and A. Zisserman, ``Very deep convolutional networks for large-scale image recognition," \emph{arXiv}:1409.1556, 2014.

\end{thebibliography}
%
% <OR> manually copy in the resultant .bbl file
% set second argument of \begin to the number of references
% (used to reserve space for the reference number labels box)

% that's all folks
\end{document}